\pdfoutput=1

\documentclass[11pt]{article}

\usepackage{ACL2023}

\usepackage{times}
\usepackage{latexsym}

\usepackage[T1]{fontenc}

\usepackage[utf8]{inputenc}

\usepackage{microtype}

\usepackage{inconsolata}

\newcommand{\rom}[1]{\uppercase\expandafter{\romannumeral #1\relax}}

\usepackage{graphicx}
\usepackage{booktabs}
\usepackage{siunitx}
\title{CoSiNES: Contrastive Siamese Network for Entity Standardization}

\author{Jiaqing Yuan$^1$, Michele Merler$^2$, Mihir Choudhury$^2$,\\ {\bf Raju Pavuluri$^2$},  {\bf Munindar P. Singh$^1$}, {\bf Maja Vukovic$^2$} \\
  $^1$ North Carolina State University, Raleigh, NC, USA \\
  $^2$ IBM Research AI, Yorktown Heights, NY, USA \\
  \texttt{\{jyuan23, mpsingh\}@ncsu.edu} \\
  \texttt{\{mimerler, choudhury, pavuluri, maja\}@us.ibm.com} \\}

\begin{document}
\maketitle
\begin{abstract} 
Entity standardization maps noisy mentions from free-form text to standard entities in a knowledge base. The unique challenge of this task relative to other entity-related tasks is the lack of surrounding context and numerous variations in the surface form of the mentions, especially when it comes to generalization across domains where labeled data is scarce. Previous research mostly focuses on developing models either heavily relying on context, or dedicated solely to a specific domain. In contrast, we propose CoSiNES, a generic and adaptable framework with Contrastive Siamese Network for Entity Standardization that effectively adapts a pretrained language model to capture the syntax and semantics of the entities in a new domain. 

We construct a new dataset in the technology domain, which contains 640 technical stack entities and 6,412 mentions collected from industrial content management systems. We demonstrate that CoSiNES yields higher accuracy and faster runtime than baselines derived from leading methods in this domain. CoSiNES also achieves competitive performance in four standard datasets from the chemistry, medicine, and biomedical domains, demonstrating its cross-domain applicability.

Code and data is available at \url{https://github.com/konveyor/tackle-container-advisor/tree/main/entity_standardizer/cosines}


\end{abstract}

\section{Introduction}


The automatic resolution of mentions in free-form text to entities in a structured knowledge base is an important task for understanding and organizing text. Two well-recognized tasks tackle entity mentions in text. \emph{Entity matching} concerns resolving data instances that refer to the same real-world entity \citep{10.14778/3421424.3421431}. The data instances usually comprise a specific schema of attributes, such as product specifications. \emph{Entity linking}, also known as entity disambiguation, associates ambiguous mentions from text with entities in a knowledge base, where precise attributes and relationships between entities are curated \citep{10.3233/SW-222986}. Both tasks involve rich context surrounding the mention and the underlying entity \citep{10.14778/3421424.3421431, 10.3233/SW-222986}. Much effort in deep learning approaches focuses on ways to leverage and encode the context surrounding mentions in text and attributes associated with entities in the knowledge base. However, little work has been done on scenarios where such rich context and precise information are not available. In domains such as finance, biology, medicine, and technology, mentions involve specialized jargon, where no context is associated with the mentions and often no attribute of the entities is available other than the mentions themselves. 

\begin{figure}[t]
\centering
\includegraphics[width=0.7\columnwidth]{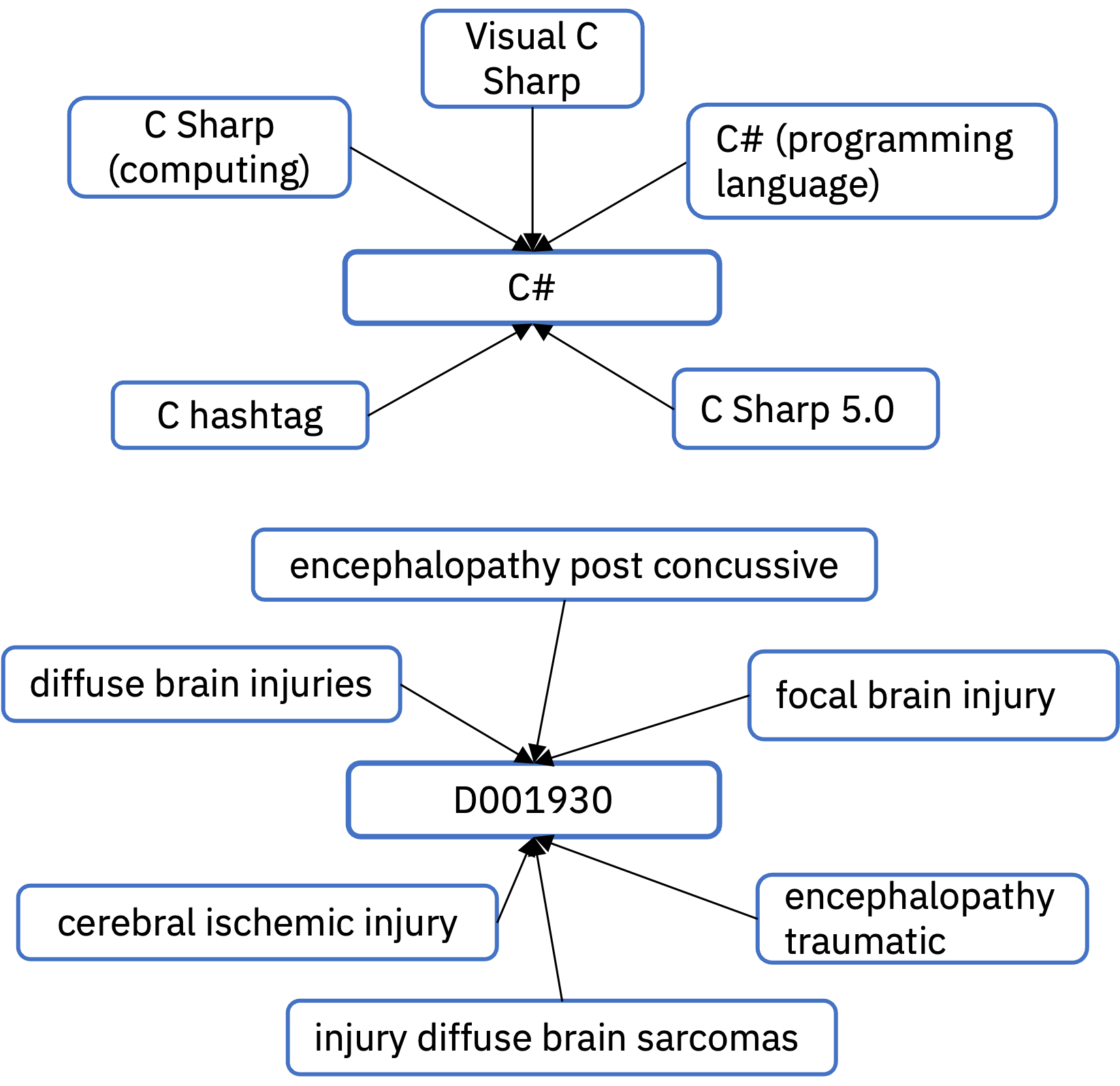}
\caption{Examples of various mentions referring to the same entity from two different domains. Top: technology, bottom: medical.}
\label{fig:examples}
\end{figure}

We tackle the challenge of missing context for entity standardization (ES) mapping, which involves mapping mentions to entities in the knowledge base across multiple domains. Due to the lack of a public dataset for ES and to foster research on the problem, we manually construct a dataset in the technology domain geared to application modernization. We propose an approach called CoSiNES for the dataset and then evaluate the generalization of CoSiNES in the biomedical domain. 

Application modernization consists in migrating legacy applications to the cloud. It relies on a faithful assessment of the technical components of such applications. Much technical information is contained in free-form textual application descriptions, but automatic extraction of such knowledge is nontrivial due to variations in how the same entities are mentioned \citep{Anup}.

Compared to the two aforementioned tasks of entity matching and linking, ES presents unique challenges. First, the mentions could have acronyms, numbers, symbols, alias, punctuation, and misspellings. Figure \ref{fig:examples} shows two examples of multiple mentions referring to the same entity. Second,  there is a lack of context surrounding the mentions, and there are no attributes or relationships for entities in the knowledge base, which the previous approaches heavily rely on. Third, large deep learning models require massive training datasets, which are not available for specialized domains. Therefore, architectures that are suited for zero-shot or few-shot learning are of great value for this task. 

Another challenge is how to perform entity standardization at scale. A naive way is to have exhaustive comparisons between each possible mention and entity pair, which is inefficient. Previous deep learning models for entity matching and entity linking usually have multiple stages \cite{papadakis-etal-2020}: first stage, such as blocking in entity matching, reduces the number of comparison pairs via a coarse-grained criterion so that the latter stages can focus on filtered candidate pairs. This multistage approach leads to globally inferior performance due to the errors accumulated along the pipeline.

We tackle these challenges with a generic framework based on Contrastive Siamese Network which efficiently adapts domain-agnostic pretrained language models (PLMs) to specific domains using a limited number of labeled examples.  Language models have shown great capacity to capture both syntactic and semantic variations of text. Our framework decouples the comparison of mention-entity pairs for training and inference so that the model can be used as a standalone encoder after training. Therefore, the embeddings of the entity from the knowledge base can be precomputed and hashed. At inference time, the running time is linear in the size of query mentions, and we can leverage existing tools, such as FAISS,\footnote{https://github.com/facebookresearch/faiss} for efficient and large-scale similarity search.

Our contributions are the following.
\begin{itemize}
    \item A generic, scalable, and adaptable framework that leverages domain-agnostic pretrained language models. 
    \item A method for generating anchored contrastive groups and a training scheme with a hybrid of batch-all and batch-hard online triplet mining. 
    \item A dataset curated for application modernization, where various mentions for technical components are manually labeled.
\end{itemize}
 We validate these contributions via comprehensive experiments with various hyperparameters, loss functions, and training schemes and show the robustness and effectiveness of the framework on our custom dataset in the technology domain. With optimal settings on our dataset, we further evaluate the framework on four datasets from the biomedical domain. We show that the framework can be adapted to other domains with minimal changes.

\begin{figure*}[ht]
\centering
\includegraphics[width=1.8\columnwidth]{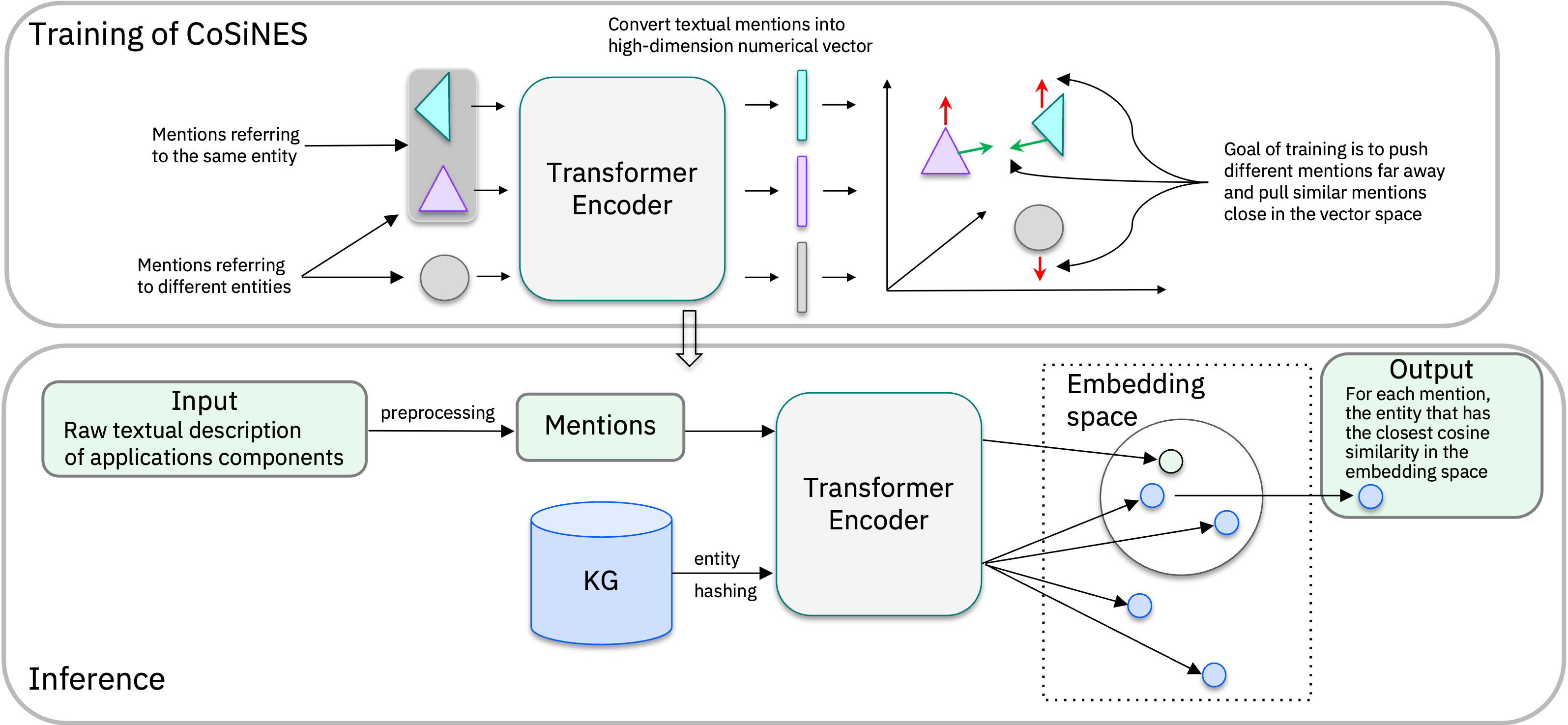}
\caption{System overview of CoSiNES.}
\label{fig:system_overview}
\end{figure*}

\section{Related Work}

Various forms of entity-related tasks have been studied by previous research, of which three are most relevant to our task.

\textbf{Entity Matching (EM)} identifies if different mentions refer to the same real-world entity, and is an important step in data cleaning and integration \citep{10.5555/2344108}. The targets of EM are records from a database, where records follow a specific schema of attributes. The goal is to find pairs of records from two databases that refer to the same entity. Whereas early approaches of EM mostly apply rule-based heuristics, recent research often relies on deep neural network \citep{10.1145/3357384.3358018, Mudgal2018DeepLF, 10.14778/3421424.3421431, 10.14778/3236187.3236198}. As the number of pairwise comparisons grows quadratically, a preprocessing step (blocking) is usually applied to reduce the number of candidate matches. The matcher then takes a pair of a mention and an entity as input and produces a probability of a match. In contrast, entity standardization comes with a predefined set of standard entities, and the mentions come with no attributes. Our method involves learning a metric function, where the model can be used as an encoder to embed mentions and entities in the same space. 

\textbf{Entity Linking (EL)} is the process of linking a mention in context with an entity in a knowledge base. Unlike entity standardization, the entities in the knowledge base, such as WikiData \citep{10.1145/2629489} and Freebase \citep{10.1145/1376616.1376746}, usually have well-structured attributes and precisely defined relationships between them. The mention comes with rich context and unstructured raw text. To leverage these two different types of contextual information, separate context-mention and graph-entity encoders are designed to produce embeddings respectively, and another neural network is used to combine and project these two embeddings to the same space \citep{Shahbazi2019EntityawareEL, yamada-etal-2022, radhakrishnan-etal-2018}. Due to the lack of context for both the mention and entity for entity standardization, we propose to use a single unified model as the encoder, which can reduce the complexity of the pipeline. 

\textbf{Entity Normalization (EN)} is widely used in the biomedical domain. The task is to map noisy mentions to entities in a well-defined reference set, such as ontologies and taxonomies \citep{C-Norm, ferre-etal-2020-handling}. The mentions usually have no context, and the entities come with no attributes, but there is a hierarchical structure in the reference set. Unlike entity standardization in the technology domain, the variations of mentions in life science are fairly standardized and synonyms are rare. The task can be well addressed with a sufficient number of training examples for each entity category, which is not the case in our setting. \citet{10.1007/978-3-030-46147-8_40} propose a similar idea using a Siamese neural network for EN. Our approach differs in the following aspects: the designed training batch-generation algorithm, the computation of the contrastive loss, and the usage of PLMs in our specialized training scheme. 

\section{Methodology}

\subsection{Problem Formulation}
We denote the set of query mentions as $\mathcal{Q} \equiv \{m_q\}$, and the set of standard entities as $\mathcal{S} \equiv \{e_s\}$. Each entity in $\mathcal{S}$ is associated with zero or more mentions referring to it $e_s \leftarrow \{m_s\}$. Importantly, there should be no overlap between the query mention set $\mathcal{Q}$ and the mentions associated with the standard entity set $\mathcal{S}$. The task is to retrieve an entity $e \in \mathcal{S}$ given $m \in \mathcal{Q}$ such that $e$ is the entity $m$ refers to. 

We tackle this task with contrastive learning by learning an embedding encoder such that mentions and entities are encoded to the same high-dimensional embedding space. The property of the embedding space is that the cosine distance between mentions of the same entity is smaller than mentions of different entities. 

We design a BERT-based Siamese neural network architecture, which acts as the embedding encoder after training. The training is conducted with a hybrid of batch-all and batch-hard online triplet mining schemes. Figure \ref{fig:system_overview} gives an overview of CoSiNES. The training (top) phase has the goal of pulling similar mentions together and pushing dissimilar mentions far away in the embedding space. After training, the inference (bottom) phase has the goal of using a Siamese neural network to project entities in the knowledge base and query mentions to the same embedding space. At inference time,  nearest neighbor search algorithms can be used to retrieve the target entity. 

\subsection{Contrastive Learning and Triplet Loss} 
Contrastive Learning  \citep{9773725, 10.1145/3561970, smith-eisner-2005} aims to group similar data points together and push dissimilar data points far apart in a high-dimensional embedding space. Equation \ref{eq:1} shows the core idea of contrastive learning. Here $x$ represents any data point in the domain, $x^+$ is a positive sample that is similar to $x$ (or from the same class as $x$), and $x^-$ is a negative sample that is dissimilar to $x$. $E$ is an encoder, which could be any neural network. And, $\textrm{dis}$ is a distance measure between the embedding vectors. 

\begin{center}
\begin{equation} \label{eq:1}
    \textrm{dis}(E(x), E(x^+)) \ll \textrm{dis}(E(x), E(x^-))
\end{equation}
\end{center}

As shown in Equation \ref{eq:2}, triplet loss is calculated based on triplets $\{x, x^+, x^-\}$, which consist of two samples from the same class and a third sample from a different class. 
The intuition 
is that the distance $\textrm{d}(x, x^-)$ 
should be larger than the distance 
$\textrm{d}(x, x^+)$ 
by a \emph{margin}. The \emph{margin} is a hyperparameter that needs to be tuned. 
\begin{center}
\begin{equation} \label{eq:2}
    \mathcal{L} = \max(\textrm{d}(x, x^+)-\textrm{d}(x, x^-)+ \text{margin}, 0)
\end{equation}
\end{center}
Based on the difference between $\textrm{d}(x, x^-)$ and $\textrm{d}(x, x^+)$, we can classify triplets into three categories: easy, semihard, and hard. See appendix \ref{appendix B} for detailed definitions. 

\subsection{Online Triplet Mining} \label{mining}
There are two different strategies of mining triplets for contrastive learning. \emph{Offline mining} generates triplets at the beginning of training. The embeddings of the whole training dataset are computed, then hard and semihard triplets are mined based on the embeddings. Offline mining is highly inefficient. First, it requires computing the embeddings for all the training data to mine the triplets. Second, as the model starts to learn, the hard and semihard triplets may turn into easy triplets. Therefore, at least for a few epochs, we need to update the triplet set frequently. 
\emph{Online triplet mining}  \citep{7298682} seeks to generate triplets on the fly within a batch.  There are two strategies to mine triplets from a batch, i.e., batch all and batch hard. We adopt the same idea in our model and propose a hybrid online mining scheme which is shown to be superior to single-mining strategy.

\subsubsection{Batch--All} To form valid triplets, a batch of training data should always include samples from more than one class, and each class should contain at least two samples. Suppose the size of the batch is $B$ and the number of all possible triplets is $B^3$. However, not all of these triplets are valid as we need to make sure each triplet comprises two distinct samples from the same class and one sample from another class. For all valid triplets in the batch, we simply select all hard and semihard triplets and compute the average loss over them. We do not include easy triplets in computing the average as it will make the loss too small. The calculations are based on the embeddings of the batch after they pass through the model. 
\subsubsection{Batch--Hard} This strategy always selects the hardest positive and negative for each anchor in the batch. Each data instance in the batch can be used as an anchor. Therefore, the number of triplets is always equal to the size of the batch.  The hardest positive has the largest $\textrm{d}(x, x^+)$ among all positives, and the hardest negative has the smallest $\textrm{d}(x, x^-)$ among all negatives. 

\subsubsection{Contrastive Group Generation} 
Based on the above discussion, a batch should include multiple samples from multiple classes. We sample batches with two steps. First, we randomly generate groups of samples from the same class with size $g$, and second, we randomly sample $b$ classes of groups to form a batch. Therefore, the effective batch size would be $B = g*b$. 

\subsection{BERT-Based Siamese Neural Network}
The canonical Siamese neural network is an architecture that consists of two towers with shared weights working in parallel on two different inputs. The outputs are passed on to a distance function to learn comparable output vectors. We extend the same idea to a batch of inputs instead of a pair of inputs. We sample the batch as described in Section \ref{mining} and feed the sampled triplets through the network. The output embeddings of the batch are used to generate valid triplets and compute the loss. The backbone of the Siamese model could be any neural network. We use the pretrained language model BERT \citep{Devlin2019BERTPO} as the backbone.



\subsection{Hashing and Retrieval}
Once the Siamese model is trained, it can be used as a standalone encoder to compute the embeddings of entities and mentions. We precompute the embeddings for all entities and save them for comparisons at inference time. For each query mention, we use the same Siamese model to get the embedding and our task is to retrieve the entity with the closest distance to the mention in the embedding space. For a query set of size $q$, we need to run the Siamese model only $q$ times, avoiding exhaustive pairwise running of the Siamese model.  Potentially, we still need to conduct a pairwise nearest neighbor search over the mention and entity embeddings. Tools such as FAISS can be leveraged to efficiently perform large-scale nearest neighbor search. 

\section{Experimental Setup}
\subsection{Dataset} 
We curate a dataset (ESAppMod) on application modernization that comprises named entities with respect to the technical stack of business applications. There are a total number of \num{640} unique entities, covering a variety of technical component categories, such as Operating System (OS), Application Server, Programming Language, Library, and Runtime. We manually extract and label \num{6412} unique mentions associated with the entities in AppMod from real application descriptions. All annotations are done by domain experts. We split the mentions \num{60}--\num{40} into train and test sets, which yields \num{3973} and \num{2439} mentions in the training and testing splits, respectively. The mentions associated with each entity are not evenly distributed, ranging from one to over a hundred. 

\subsection{Hyperparameter Tuning}
Implementing our framework involves many design choices and hyperparameters. To facilitate performance at scale, the tradeoff between accuracy and inference time is crucial. We experimented with different sizes of BERT as the backbone of CoSiNES, including BERT-tiny, BERT-mini, BERT-small, BERT-medium, and BERT-base. For triplet mining, we evaluated batch--all, batch--hard, and a hybrid of the two. For the measure of distance, we investigated cosine, Euclidean, and squared Euclidean distance. For the hyperparameters, we evaluated different values of margin, learning rate, and batch size detailed in appendix \ref{appendix C}. All training experiments were carried out on an NVIDIA A100 GPU with 40GB memory. We use the tool Ray.tune\footnote{https://docs.ray.io/en/latest/tune/index.html} for hyperparameter tuning. Inference times were computed as the cumulative time to predict all 2,439 mentions in the test set on the CPU of Macbook pro with 2.3 GHz Quad-Core Intel Core i7, 32 GB 3733 MHz LPDDR4X RAM. We report the median inference time of 10 runs. 

\subsection{Baselines}
We compare CoSiNES with four baselines.

\textbf{TF-IDF} A model that computes TF-IDF embeddings learned from training data\citep{Anup}. 

\textbf{GNN} A graph neural network that treats each entity or mention as a chain. Each character represents a node in the graph and its embedding representation is learned during training. The average of the character embeddings are used to represent entity names and mentions \cite{FAN2022778}. 

\textbf{BERT} We use the mean of last layer outputs of all tokens from BERT\_small \cite{bhargava2021generalization} to represent entities and mentions. This is the same backbone used to train CoSiNES.

\textbf{GPT3\footnote{https://beta.openai.com/docs/guides/embeddings/}} We use the embedding GPT-3 api from OpenAI to compute the embeddings using model \texttt{embedding-ada-002}.

\begin{table}[t]
\centering
\begin{tabular}{l@{~} c c c r}
  \toprule
  Model & T@1 & T@3 & T@5 & Inf. Time\\ 
 \midrule
 TF-IDF & 69.94 & 85.36 & 88.44 & 60 \\
 GNN  & 67.20 & 79.29 & 82.49 & 29 \\
 BERT  & 32.64 & 47.23 & 54.82 & 17 \\
 GPT3  & 77.24 & \textbf{90.24} & \textbf{93.56} & 240 \\
 CoSiNES & \textbf{80.40} & 88.68 & 90.98 & 11\\ 
 \bottomrule
\end{tabular}
\caption{\label{table:results}Experimental results on ESAppMod. T@1: top-1 retrieval accuracy. Inf.\ Time refers to total inference time in seconds.}
\end{table}

\section{Results and Discussions}
Table \ref{table:results} shows the comparative results on our dataset. Our model outperforms all baselines by a significant margin in terms of top--1 retrieval accuracy: 10.46\% over TF-IDF, 13.2\% over GNN, 47.76\% over BERT, and 3.16\% over GPT3. Through comprehensive experimentation, we observe that the best performance model has the BERT-small as the backbone. The learning rate is set to \num{1e-5}, contrastive group size is \num{10}, and the batch size of groups is \num{16}, which makes the effective batch size \num{160}. We set the margin to \num{2}. 

\begin{figure}[ht]
\centering
\includegraphics[width=0.95\columnwidth]{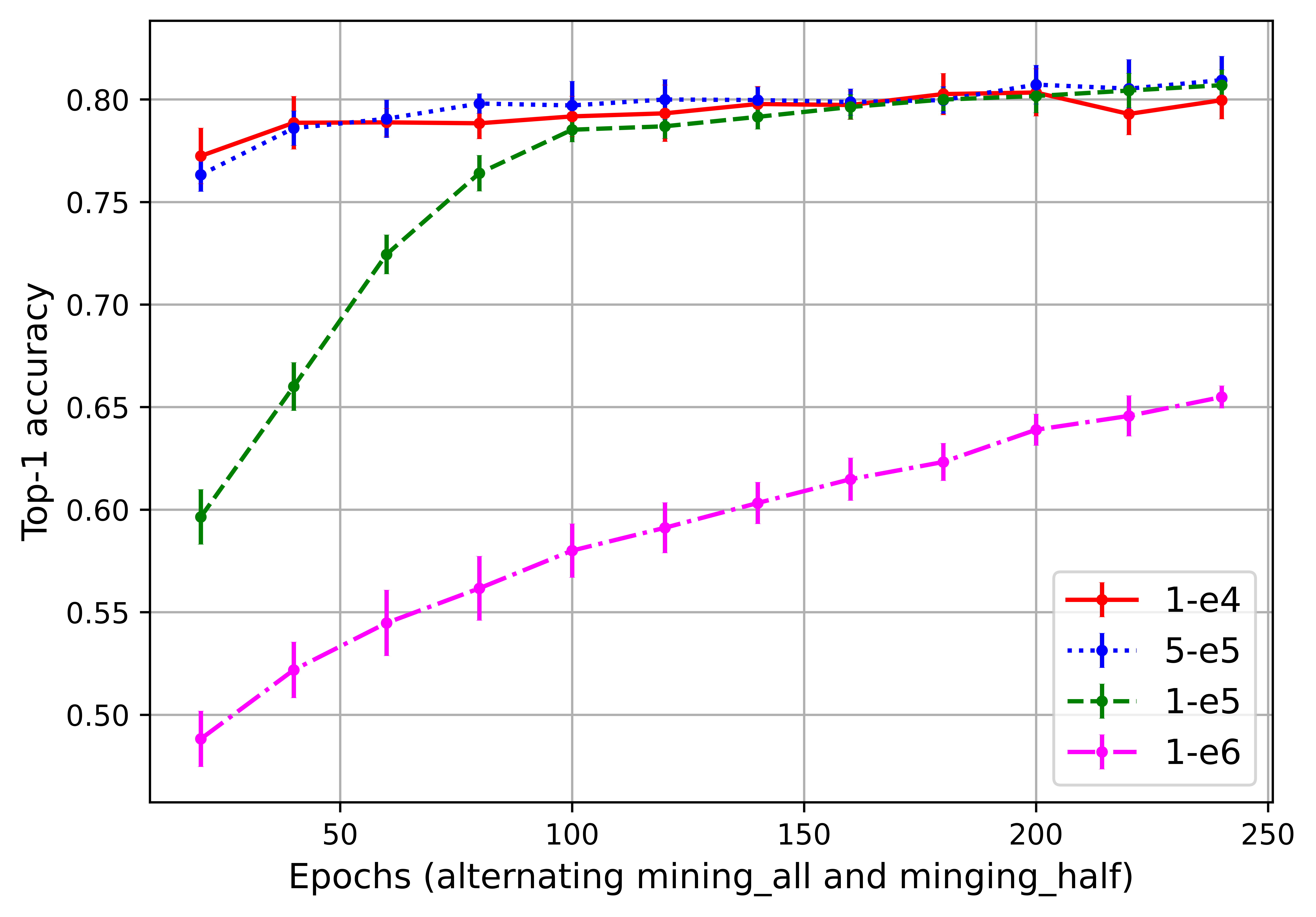}
\caption{Five--fold cross--validation with different learning rates on training data.}
\label{fig:different-learning-rate}
\end{figure}

\subsection{Learning Rate}
To investigate how different learning rates affect the convergence of the Siamese model on our dataset, we run five-fold cross-validation with four learning rates (\num{1e-4}, \num{5e-5},  \num{1e-5}, and \num{1e-6}) on the training data, as shown in Figure \ref{fig:different-learning-rate}. For each learning rate, we experiment with different numbers of epochs, ranging from \num{10} to \num{200} with an interval of \num{10}. The X axis is the number of epochs for each experiment and the Y axis is the top--1 accuracy. The average of the five-fold top--1 accuracy is shown for each dot in the figure, together with the standard deviation across five folds. As we can see, the learning rate affects how fast and stably the model converges, and most of them reach similar performance when trained for enough number of epochs. This indicates that the Siamese model is robust with respect to the learning rate. 
We set the learning rate to be 1e-5 as it tends to have a smaller deviation of performance.

\subsection{Hybrid Triplet Mining}
\begin{figure*}[ht]
\centering
\includegraphics[width=2\columnwidth]{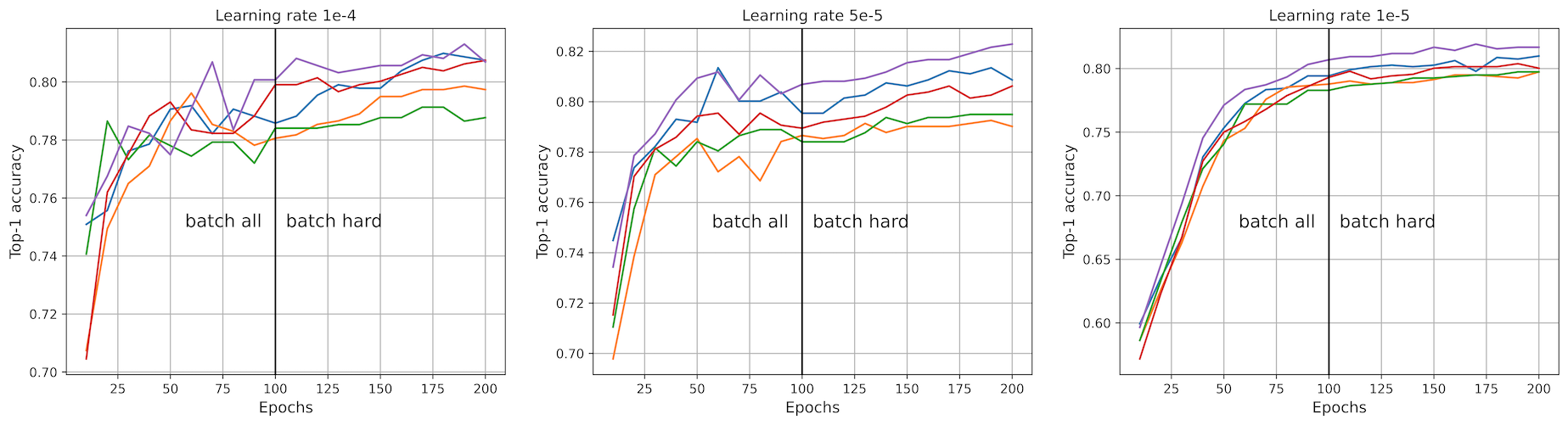}
\caption{Hybrid triplet mining with different learning rates for five-fold cross validation.}
\label{fig:hybrid mining}
\end{figure*}

We propose a hybrid of batch--all and batch--hard triplet mining during training. Figure \ref{fig:hybrid mining} shows the training process with \num{200} epochs with the above three learning rates, of which the first \num{100} epochs apply batch--all triplet sampling and the second \num{100} epochs employ batch--hard triplet sampling. The result shows that for the first batch--all \num{100} epochs, the training of \num{1e-4} and \num{5e-5} is unstable and performance oscillates greatly. When batch--hard mining comes into play, the training becomes much smoother and the performance continues to improve steadily for all three learning rates. This experiment shows that the hybrid mining scheme improves the top--1 accuracy by around 2\% compared to the single-mining strategy.

\subsection{Model Size}
\begin{figure}[ht]
\centering
\includegraphics[width=0.9\columnwidth]{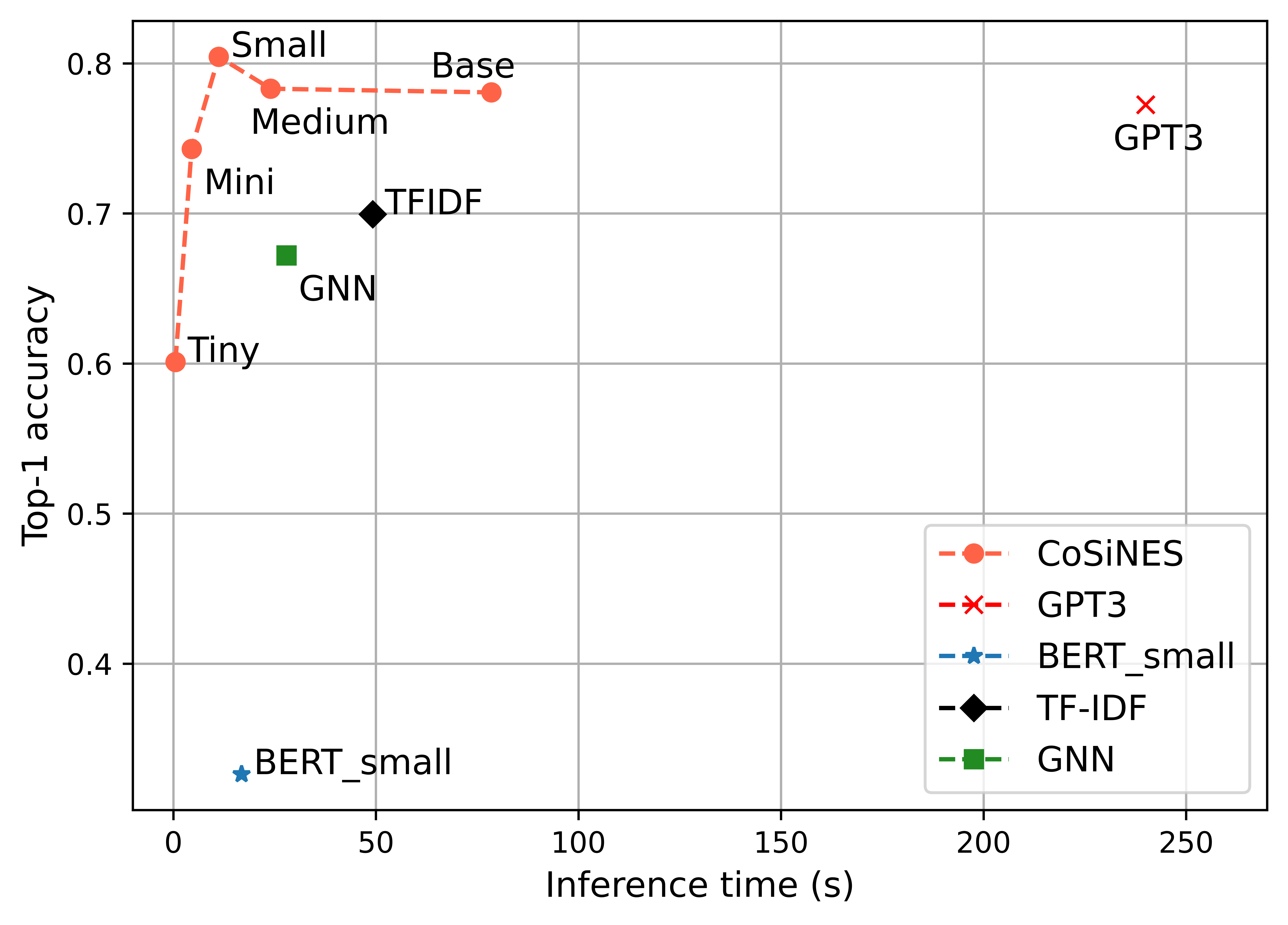}
\caption{Accuracy versus efficiency between the proposed models on the ESAppMod dataset. The CoSiNES line represents  different size of BERT as backbone. }
\label{fig:different-sizes-BERT}
\end{figure}

Normally, there is a tradeoff between model accuracy and efficiency. 
Therefore, we experiment with different sizes of BERT as backbone to find a balance between performance and running time. Figure \ref{fig:different-sizes-BERT} shows the inference time on the testing set with top--1 accuracy. The results show that CoSiNES with BERT-small achieves the best performance and fast inference time. Although the GPT3 embeddings achieve  performance close to CoSiNES, running inference using the GPT3 OpenAI api is inefficient. 

\begin{figure}[ht]
\centering
\includegraphics[width=0.95\columnwidth]{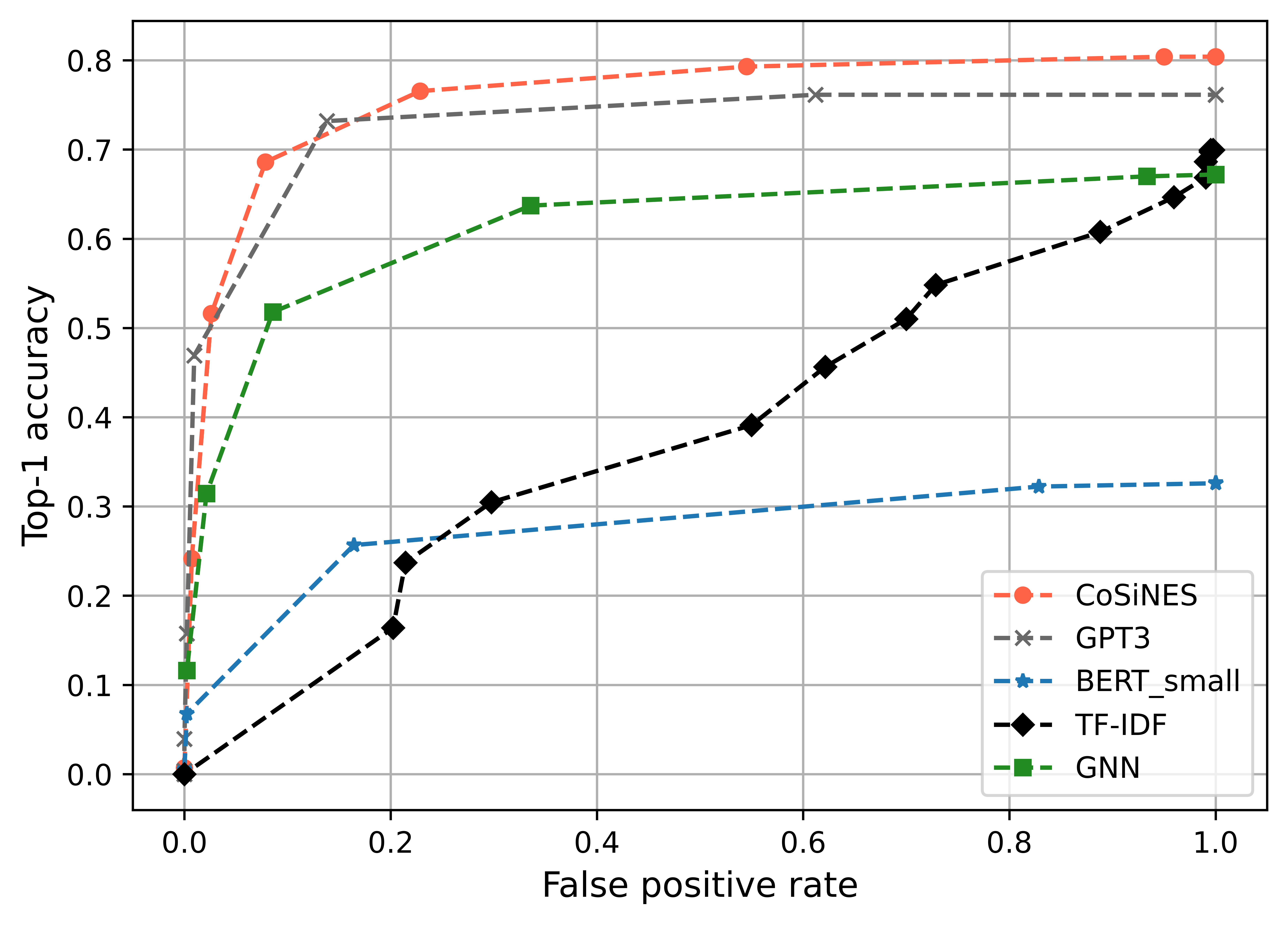}
\caption{ROC Curves on the ESAppMod dataset.}
\label{fig:roc}
\end{figure}

\subsection{ROC Curve}
For a comprehensive comparison between our model and the baselines, we conduct an experiment to compute the receiver operating characteristic (ROC) curve. We add \num{420} previously unseen relevant but negative mentions from the technology domain that do not refer to any entities in the training set, and calculate the false positive rate under different thresholds. Figure \ref{fig:roc} shows that our proposed model has a larger area under the curve, which demonstrates its superior performance over the baselines. 

\subsection{Qualitative Error Analysis}
We examine the predictions from CoSiNES on ESAppMod and categorize the following error types. Table \ref{table:error examples} shows a few examples for each of these types. 

\textbf{Misspelling}. When a mention has an error in the spelling, the tokens returned by PLMs could be very different, which leads to mismatch. This is a challenge for PLMs, whereas human could easily handle, e.g. ``Andriod" vs ``Android".

\textbf{Acronym}. Linking acronyms to full expressions seem to be a trivial task for humans, however, CoSiNES falls short of this capability. The rescue might be to design a task specialized for recognizing acronyms for PLMs.

\textbf{Multi-match}. This is the most common error where multiple entities partially match with the mention in the surface form. One way to address this issue is to enrich the training dataset with various mentions, which is not always within easy reach. Another potential approach is to integrate external knowledge about entities so that the model can refer to. 

\textbf{No-match}. When the entity and mention have no match at all in the surface form, it is unlikely for the model to retrieve the correct target, especially no context can be leveraged. Therefore, external knowledge could be particularly useful in this case.  

\begin{table*}[t]
\scriptsize

\centering
\begin{tabular}{l@{~} l l l l l l}
  \toprule
  Error type  & Mention & Target entity & Top-5 retrieved entities\\ 
 \midrule
 Misspelling & Andriod &  Android &IBM ILOG Views /  Oracle Real-Time Decisions (RTD) /  BeOS / Ingres / etcd\\
 & Visusal Basic &  Visual Basic & Clarify|Clear Basic /  BASIC /  IBM Basic Assembly Language / Pervasive PSQL / ADABAS\\
 \midrule

 Acronym  & NES &  Netscape Enterprise Server & Mobile / SAS /  iOS /  Powershell /  MinIO\\
    & IIB &  IBM Integration Bus & Visual Basic / VB.NET / Clarify|Clear Basic / IIS|* / Ada\\
 \midrule
 Multi-match & Cordova Android &  Apache Cordova & Android	/ Apache Cordova / Cisco IOS / Perl|Oraperl / Keycloak\\
    & MQ 9.1 & IBM Websphere MQ & Microsoft MQ /  MQ Client /  IBM Websphere MQ /  Qiskit /  IBM WebSphere MQ Telemetry \\
    & Open Liberty & WebSphere Liberty & OpenROAD / WebSphere Liberty / Virtual Appliance /\\
    & & & OpenVPN / Microsoft System Center Endpoint Protection\\
 \midrule
 No-match & AS400 &  IBM Power Systems& DB400 / Asterisk / Primavera P6 / EAServer / Microsoft Excel\\
       &EAP & JBoss & XAMPP / F5 Secure Web Gateway Services / Java|Java Web Start / \\
       &&&UltiDev Web Server Pro (UWS) / A-Auto Job Scheduling Software\\
 \bottomrule
\end{tabular}
\caption{\label{table:error examples} Examples for each type of errors on ESAppMod.}
\end{table*}

\section{Adaptation to Biomedical Domain}
 We show how to adapt our framework to the biomedical domain with minimal changes. 

\subsection{Datasets}
We consider four public datasets, ncbi, bc5cdr-disease, bc5cdr-chemical, and bc2gm, covering three types of entities: chemicals, diseases, and genes. Details and statistics regarding the datasets can be found in apprendix \ref{appendix A}.

\subsection{Baselines}
We compare our framework with three models. 

\textbf{TF-IDF} Like the baseline for ESAppMod, we implement a straightforward TF-IDF model \cite{Anup} based on the knowledge database for each dataset and apply nearest-neighbor search for testing. 

\textbf{BioBERT ranking}
Use BioBERT \citep{10.1093/bioinformatics/btz682} to encode concepts and mentions without fine-tuning. BioBERT is a large biomedical language representation model pretrained with PubMed abstracts and PMC full-text articles. 

\textbf{BioSyn} BioSyn \citep{Sung2020BiomedicalER} is the state-of-the-art model for biomedical entity normalization with synonym marginalization and iterative candidate retrieval. The model leverages sparse embedding from TF-IDF and dense embedding from BioBERT. 

\subsection{Domain Adaptation}
For domain adaptation, it would be ideal if we can make none or a few changes to the model architecture and training process. Therefore, we follow all experimental settings, such as learning rate, margin, contrastive group generation, and hybrid training scheme from the experiments on our proposed datasets. The most significant change is that to adapt to a new domain, we use dmis-lab/biobert-v1.1\footnote{https://huggingface.co/dmis-lab/biobert-v1.1} in replacement of the regular BERT as our backbone. We conduct all experiments on two NVIDIA A100 GPUs and adjust the batch size for each dataset based on the lengths of the mentions. 

\subsection{Results}
The results are shown in Table \ref{table:biomedical_results}. We reproduce the BioBERT experiment reported by \citep{tutubalina-etal-2020-fair} using the embedding of the [CLS] token as the representation. The results are almost identical. The minor differences might be due to different versions of the pretrained language model. 

The performance of BioSyn reported by \citet{Sung2020BiomedicalER} is high. However, as pointed out by \citet{tutubalina-etal-2020-fair}, the original testing splits used by \citet{Sung2020BiomedicalER} have significant overlapping mentions with the knowledge base. Therefore, \citeauthor{tutubalina-etal-2020-fair} removed all the duplicates and produced refined testing splits. We follow the performance of BioSyn reported by them. 

The results show that CoSiNES significantly outperforms the baselines of TF-IDF and BioBERT ranking in terms of top-k accuracy. 
CoSiNES achieves competitive results with BioSyn on all the datasets. Given that we didn't change any hyperparameters or architectures of CoSiNES, and directly applied the framework to new domains, we demonstrate the cross-domain applicability of CoSiNES.

\begin{table}[t]
\centering
{\small
\begin{tabular}{l@{~} r r r r r r}
\toprule
   & ncbi & bc5cdr-d & bc5cdr-c & bc2gm \\
 \midrule
 \textbf{TF-IDF@1} & 59.31 & 61.34 & 71.76 & 67.01\\ 
 TF-IDF@3 & 69.61 & 69.41 & 76.24 & 76.55\\ 
 TF-IDF@5 & 74.02 & 73.21 & 78.59 & 79.90\\ 
 \midrule
 \textbf{BioBERT@1} & 47.55 & 64.23 & 79.55 & 68.12\\ 
 BioBERT@3 & 57.35 & 74.89 & 81.65 & 74.11\\ 
 BioBERT@5 & 61.77 & 79.45 & 82.82 & 76.04\\ 
 \midrule
 \textbf{BioSyn@1} & 72.5 & \textbf{74.1} & \textbf{83.8} & \textbf{85.8}\\ 
 BioSyn@3 & - & - & - & -\\ 
 BioSyn@5 & - & - & - & -\\ 
 \midrule
 \textbf{CoSiNES@1} & \textbf{72.55} & 73.52 & 81.65 & \textbf{85.79}\\ 
 CoSiNES@3 & 80.39 & 78.39 & 85.88 & 90.66\\ 
 CoSiNES@5 & 81.37 & 80.52 & 87.76 & 91.68\\ 
 \bottomrule
\end{tabular}
}
\caption{Results on four datasets from the biomedical domain. @1: top-1 accuracy. Here, bc5cdr-d means bc5cdr-disease and bc5cdr-c means bc5cdr-chemical.}
\label{table:biomedical_results}
\end{table}

\section{Conclusion}
We propose a generic, scalable, and adaptable framework CoSiNES for the entity standardization task, which maps various mentions to standard entities in the knowledge base. We first construct a new dataset ESAppMod in the technology domain and demonstrate the superiority of our framework over four other models. We conduct comprehensive experiments regarding batch size, learning rate, margin, loss calculation and different sizes of BERT, with our designed contrastive group generation and hybrid triplet mining, and show that the framework is rather robust with respect to hyper-parameters. With the optimal setting on our dataset, we further show that our model can be easily adapted to new domains with minimal changes by achieving competitive performance on four benchmark datasets from the biomedical domain covering three different types of entities. 

After examining the errors produced by the framework on our proposed dataset, we categorize four different types of errors and defer to future work with the following directions: (1) integrating the framework with external knowledge. For multi-match errors, where multiple entities partially match with the mention, it would be ambiguous to retrieve the target entity. For no-match errors, external knowledge could provide extra information; (2) Adversarial training for misspellings. For technical terms, misspelling could lead to completely different tokenization of the mentions; (3) Construct new or augment the existing training dataset with acronym samples. The pretrained language models are not specialized in recognizing acronyms. Therefore, it would be worthwhile endowing PLMs with such capability. 

\section*{Limitations}
We focuses on resolving various mentions from different domains. Although we have tested our framework on multiple datasets, it relies on a human-annotated dataset and effort should be taken to investigate how the model performs with emerging domains without human-annotated data. Our model works with mentions that have been extracted from raw text. It would be more practical if the model could work with raw text directly and interact with another mention-extraction module. The performance of the model is largely affected by the surface form of the mentions, although our framework is robust to variations in the surface form, it would be more beneficial to further investigate how adversarial turbulence in the mentions could affect the behaviors of the framework. 

\section*{Ethics Statement}
The domain and data we work with don't involve any personal information and are all publicly available. However, as the work could be potentially applied in the medical domain to resolve mentions of disease, discretion is advised when any medical decisions or diagnostics are made with the assistance of the model. 


\bibliography{anthology,jiaqing}
\bibliographystyle{acl_natbib}

\appendix

\section{Biomedical Datasets Descriptions and Statistics}
\label{appendix A}
Detailed descriptions of the datasets can also be found in \citet{tutubalina-etal-2020} and \citet{Sung2020BiomedicalER}.

\textbf{NCBI Disease Corpus} 
NCBI Disease Corpus \citep{Dogan2014NCBIDC}  contains manually annotated disease mentions extracted from 793 PubMed abstracts and their corresponding concepts in the MEDIC dictionary \citep{Davis2012MEDICAP}. The July 6, 2012 version of MEDIC has \num{11915} CUIs (concept ids) and \num{71923} synonyms (mentions).

\textbf{BioCreative \rom{5} CDR}
BioCreative V CDR (BC5CDR) \citep{Li2016BioCreativeVC} is a challenge for extracting chemical-disease relations. There are manual annotations for both chemical and disease from \num{1500} PubMed abstracts. Like the NCBI disease corpus, disease mentions are mapped into the MEDIC dictionary. The chemical mentions are mapped into the Comparative Toxicogenomics DataBase (CTD) \citep{Davis2018TheCT}. The Nov 4, 2019 version of CTD contains \num{171203} CUIs and \num{407247} synonyms. 
 
\textbf{BioCreative \rom{2} GN}
BioCreative \rom{2} GN (BC2GN) \citep{Morgan2008OverviewOB} contains human gene and gene product mentions from PubMed abstracts. It has \num{61646} CUIs and \num{277944} synonyms \citep{tutubalina-etal-2020-fair}.

\begin{table}[h]
\centering
\begin{tabular}{l@{~} r r r r r}\toprule
   & KG entity &  KG mention & Test mention\\
 \midrule
ncbi & \num{12554} & \num{73024} & \num{204} \\
bc5cdr-d  & \num{12511} & \num{73126} & \num{657}\\
 \midrule
bc5cdr-c  & \num{171284} & \num{407600} & \num{425}\\
bc2gm  & \num{67370} & \num{277944} & \num{985}\\
\bottomrule
\end{tabular}
\caption{Diomedical datasets statistics. Here, KG means knowledge base, bc5cdr-d means bc5cdr-disease and bc5cdr-c means bc5cdr-chemical.}
\label{table:biomedical_stats}
\end{table}

\section{Triplet Types}
\label{appendix B}
As shown in Equation \ref{eq:3}, triplet loss is calculated based on triplets $\{x, x^+, x^-\}$, which always consist of two samples from the same class and a third sample from a different class. We usually call $x$ the anchor of the triplet, $x^+$ the positive sample, and $x^-$ the negative sample. The intuition behind the loss function is that the distance $\textrm{d}(x, x^-)$ between the anchor and negative should be larger than the distance $\textrm{d}(x, x^+)$ between the anchor and positive by a \emph{margin}. The \emph{margin} is a hyperparameter that needs to be tuned. 
\begin{center}
\begin{equation} \label{eq:3}
    \mathcal{L} = \max(\textrm{d}(x, x^+)-\textrm{d}(x, x^-)+ \text{margin}, 0)
\end{equation}
\end{center}
Based on the difference between $\textrm{d}(x, x^-)$ and $\textrm{d}(x, x^+)$, we can classify triplets into three categories. 
\begin{figure}[htbp]
\centering
\includegraphics[width=0.8\columnwidth]{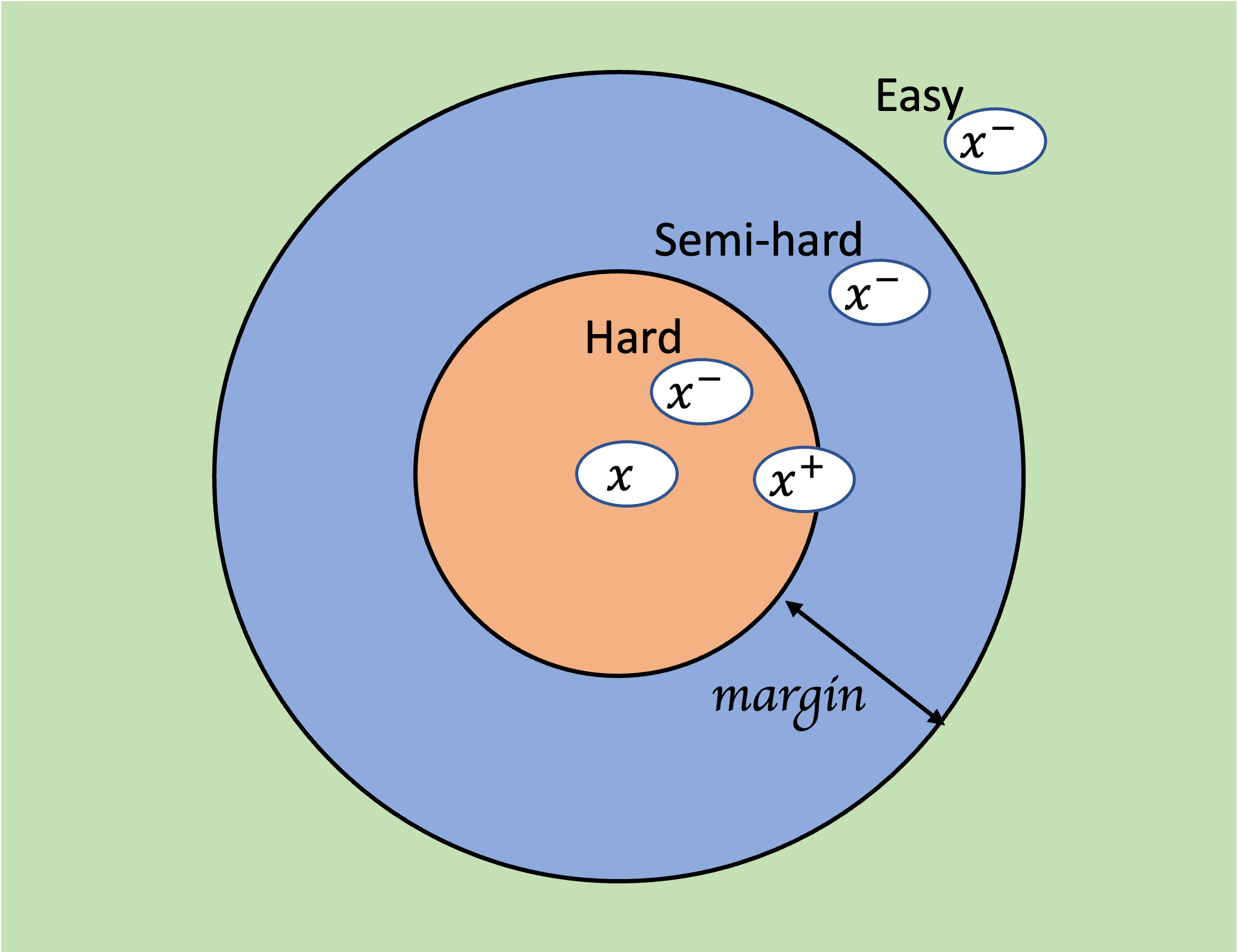}
\caption{Different types of triplet samples.} 
\label{fig:triplet}
\end{figure}
\begin{itemize}
    \item Easy triplets, which have a loss of zero based on Equation \ref{eq:2}. Therefore, easy triplets provide no learning signal to the model.   
        $$\textrm{d}(x, x^-) - \textrm{d}(x, x^+) > \text{margin}$$
    \item Semihard triplets, which have a loss less than the \emph{margin}. 
    $$ 0 < \textrm{d}(x, x^-) - \textrm{d}(x, x^+) < \text{margin}$$
    \item Hard triplets, which are most informative for the model. 
    $$\textrm{d}(x, x^-) - \textrm{d}(x, x^+) < 0$$
\end{itemize}

\section{Hyperparameter Search}
\label{appendix C}
We have done the following hyperparameter search grid on ESAppMod

\begin{table}[h]
\centering
\begin{tabular}{l@{~} l}
\toprule
Batch Size & \num{4}, \num{8},  \num{16}, \num{32} \\
Learning Rate & \num{1e-3}, \num{1e-4},  \num{1e-5}, \num{1e-6}\\
Margin  & \num{0.5}, \num{1},  \num{2}, \num{5}, \num{10}\\
\bottomrule
\end{tabular}
\caption{Hyperparameter search on ESAppMod}
\label{table:hyperparameter}
\end{table}

\end{document}